\documentclass{article}

\usepackage[preprint]{neurips_2024}

\usepackage[utf8]{inputenc} 
\usepackage[T1]{fontenc}    
\usepackage{hyperref}       
\usepackage{url}            
\usepackage{booktabs}       
\usepackage{amsfonts}       
\usepackage{graphicx}       
\usepackage{nicefrac}       
\usepackage{microtype}      
\usepackage{xcolor}         
\usepackage{listings}
\usepackage{tikz}
\usepackage{pgfplots}
\usepackage{multirow}
\usepackage{subcaption}
\pgfplotsset{compat=1.18}
\usepackage{colortbl}
\definecolor{best}{RGB}{225,235,255} 
\definecolor{darkgreen}{RGB}{0,120,60}
\definecolor{darkred}{RGB}{180,40,40}
\usepackage{siunitx}
\lstdefinestyle{code}{
    basicstyle=\ttfamily\small,
    keywordstyle=\color{blue},
    commentstyle=\color{gray},
    stringstyle=\color{green!50!black},
    breaklines=true,
    frame=single,
    columns=fullflexible,
    keepspaces=true,
    showstringspaces=false
}

\title{Exploring different approaches to customize language models for domain-specific text-to-code generation}

\author{%
  Luís Freire\thanks{This work builds upon research originally conducted as part of Luis Freire’s master’s thesis~\cite{freire2025}, which contains additional methodological details and extended experimental analyses.}
  \\
  Technical University of Denmark \\
  \texttt{luis.freire@live.com.pt} \\
  \And
  Fernanda A. Andaló \\
  The LEGO Group \\
  \texttt{fernanda.andalo@LEGO.com} \\
  \AND
  Nicki Skafte Detlefsen \\
  Technical University of Denmark \\
  \texttt{nsde@dtu.dk} \\
}

\begin{document}

\maketitle

\begin{abstract}
Large language models (LLMs) have demonstrated strong capabilities in generating executable code from natural language descriptions. However, general-purpose models often struggle in specialized programming contexts where domain-specific libraries, APIs, or conventions must be used. Customizing smaller open-source models offers a cost-effective alternative to relying on large proprietary systems.
In this work, we investigate how smaller language models can be adapted for domain-specific code generation using synthetic datasets. We construct datasets of programming exercises across three domains within the Python ecosystem: general Python programming, Scikit-learn machine learning workflows, and OpenCV-based computer vision tasks. Using these datasets, we evaluate three customization strategies: few-shot prompting, retrieval-augmented generation (RAG), and parameter-efficient fine-tuning using Low-Rank Adaptation (LoRA).
Performance is evaluated using both benchmark-based metrics and similarity-based metrics that measure alignment with domain-specific code. Our results show that prompting-based approaches such as few-shot learning and RAG can improve domain relevance in a cost-effective manner, although their impact on benchmark accuracy is limited. In contrast, LoRA-based fine-tuning consistently achieves higher accuracy and stronger domain alignment across most tasks. These findings highlight practical trade-offs between flexibility, computational cost, and performance when adapting smaller language models for specialized programming tasks.
\end{abstract}

\section{Introduction}
\label{sec:introduction}

Large Language Models (LLMs) have significantly advanced software development workflows. Recent models generate code from natural language descriptions and assist with tasks such as code completion, debugging, and documentation generation~\cite{minaee2025}. These systems translate natural language specifications into executable code.

Despite these advances, most code-generation models are trained as general-purpose systems that cover many programming languages and tasks. Although this enables strong overall performance, it can limit effectiveness in specialized domains where code must follow specific libraries, frameworks, or conventions. As a result, models may produce syntactically correct solutions that are misaligned with the intended APIs or domain constraints.

Another practical challenge is the cost and deployability of frontier models. Large proprietary systems often require substantial computational resources~\cite{kaplan2020} and are typically accessible only through external APIs. This limits their use in environments where data privacy, deployment constraints, or computational cost are critical. Consequently, there is growing interest in adapting smaller open-source language models that can be customized and deployed locally while maintaining competitive performance.

To address these challenges, we investigate how smaller open-source language models can be customized for domain-specific code generation. We compare three adaptation strategies and analyze their impact on performance, computational cost, and domain alignment in specialized programming contexts, as well as practical factors such as implementation complexity and data requirements.

Customizing a pre-trained language model for a specialized domain can be achieved through several adaptation strategies that differ in complexity, computational requirements, and the extent to which they modify model parameters. In this work, we focus on three complementary approaches: few-shot learning, which guides generation through in-context examples without updating model parameters~\cite{brown2020}; Retrieval-Augmented Generation (RAG), which augments prompts with relevant examples retrieved from an external dataset~\cite{gao2024}; and fine-tuning, where the model is specialized using domain-specific data through parameter-efficient techniques such as Low-Rank Adaptation (LoRA)~\cite{han2024}.

Despite the growing number of language model adaptation techniques, there is limited empirical understanding of how they compare for domain-specific code generation with smaller open-source models. In particular, the trade-offs between prompting, retrieval-based methods, and parameter-efficient fine-tuning remain unclear when models must generate code that follows specialized APIs and library conventions.

To study these trade-offs, we evaluate customization across three programming domains within the Python ecosystem: general Python programming exercises, machine learning workflows using Scikit-learn~\cite{scikit-learn}, and computer vision tasks using OpenCV~\cite{opencv_library}. These domains represent distinct programming contexts within the same language while requiring different types of domain knowledge. General Python tasks emphasize logical reasoning and language fundamentals, whereas Scikit-learn and OpenCV tasks rely on specialized APIs and library conventions. Keeping the programming language constant while varying the domain allows us to isolate the effects of model adaptation on domain-specific code generation.

Another challenge in domain-specific code generation is the availability of suitable training data. High-quality datasets pairing natural language descriptions with correct code implementations are scarce, particularly for specialized programming tasks. To address this limitation, recent work has explored synthetic data generation, where LLMs produce training examples for smaller models. This strategy can be viewed as a form of knowledge distillation.

Using this framework, we evaluate two open-source code-generation models—StarCoder~\cite{li2023} and DeepSeekCoder~\cite{guo2024}—and analyze performance changes under different adaptation strategies.

We evaluate these approaches using a framework combining benchmark-based and similarity metrics. Benchmark evaluations measure the functional correctness of generated solutions through automated test cases, while similarity metrics capture how closely generated code aligns with domain-specific coding conventions.

This work makes the following contributions:
\begin{enumerate}
    \item A systematic empirical comparison of prompting, retrieval-based generation, and parameter-efficient fine-tuning for domain-specific code generation.
    \item A pipeline for generating synthetic datasets of domain-specific programming tasks.
    \item An evaluation framework combining functional correctness and domain alignment metrics.
    \item Empirical insights into the specialization of small code-generation models.
\end{enumerate}

Together, the contributions provide practical guidance for adapting smaller language models to specialized programming tasks while balancing performance, cost, and deployment constraints.

\section{Background}

LLMs are typically trained with autoregressive objectives that predict the next token from the preceding context. Most modern models rely on the Transformer architecture~\cite{vaswani2017}, which uses self-attention to capture dependencies between tokens. Trained on large corpora of text, these models can learn patterns in programming syntax and code structure, enabling tasks such as code completion and program synthesis~\cite{chen2021}.

During inference, LLMs can perform \emph{in-context learning}, where task behavior is inferred from examples provided in the prompt without updating model parameters. In this setting, prompt examples act as demonstrations from which the model generalizes input–output mappings to new queries. The effectiveness of this process depends on factors such as the number and ordering of examples, the clarity of demonstrations, and the available context length.

A key limitation is the fixed-size \emph{context window}, which determines the maximum number of tokens the model can process at once. All prompt instructions and previously generated tokens must fit within this window. As a result, prompt-based approaches such as few-shot learning~\cite{brown2020} can only include a limited number of examples to guide the model.

Prompt-based methods can be extended by incorporating external information sources. Retrieval-Augmented Generation (RAG) retrieves relevant documents or examples from an external dataset at inference time~\cite{gao2024}. This approach combines parametric knowledge with non-parametric external memory, allowing models to access domain-specific information not encoded in their weights. However, the retrieved content must still fit within the model’s context window, which limits how much external information can be incorporated into the prompt.

Another approach to adapting language models is \emph{fine-tuning}, which modifies the model parameters to incorporate new knowledge directly. However, updating all parameters can be computationally expensive for modern LLMs. Parameter-efficient fine-tuning techniques address this challenge by introducing small trainable components while keeping most parameters frozen~\cite{han2024}. A widely used method is Low-Rank Adaptation (LoRA), which adds trainable low-rank matrices to selected network layers to enable efficient task specialization~\cite{hu2022}.

A practical challenge in domain-specific code generation is the limited availability of suitable training data. Many programming domains lack datasets pairing natural language descriptions with corresponding code implementations. Recent work addresses this limitation using synthetic datasets generated by large language models, where a larger teacher model produces examples for training smaller models. This process can be viewed as a form of knowledge distillation~\cite{hinton2015} and has been successfully applied in studies such as TinyStories~\cite{eldan2023} and Phi-1~\cite{gunasekar2023}.

Evaluating code generation models presents challenges beyond those in natural language generation. Benchmark-based evaluation measures functional correctness by executing generated programs against predefined test cases, as in benchmarks such as HumanEval~\cite{chen2021}, but may overlook differences in coding style, structure, or API usage that are important in domain-specific settings. Similarity-based metrics instead measure how closely generated code aligns with reference implementations. Traditional metrics such as BLEU~\cite{papineni2002} and ROUGE~\cite{lin2004} rely on n-gram overlap and often fail to capture the structural and semantic properties of source code, while metrics such as CodeBLEU~\cite{ren2020} incorporate syntax-tree and data-flow information. Because functional correctness and structural similarity capture complementary aspects of code quality, we combine both evaluation strategies in our experiments.

\section{Overview of the customization pipeline}

\begin{figure}[h]
\centering
\includegraphics[width=0.7\linewidth]{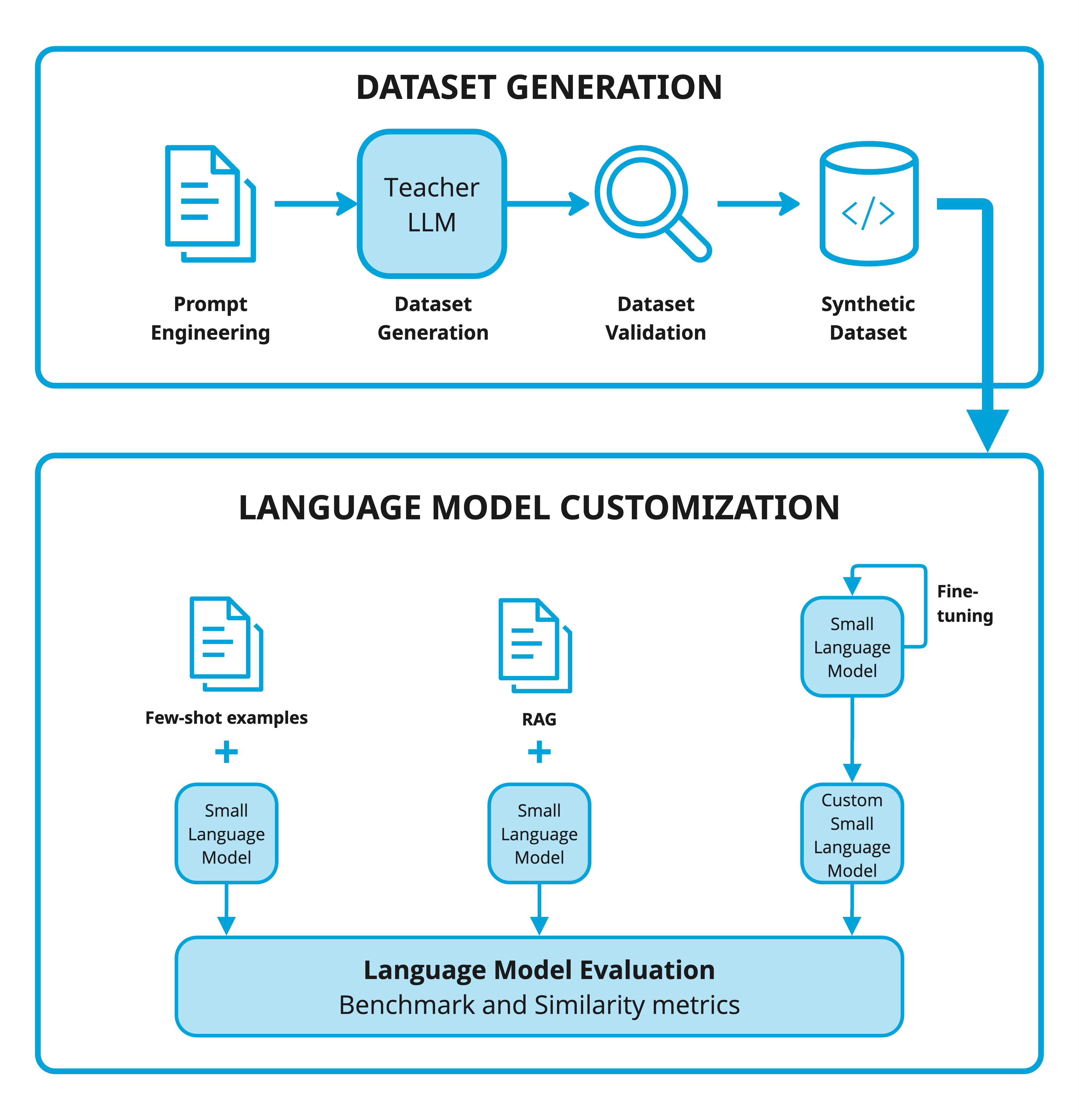}
\caption{Overview of the customization pipeline. Synthetic programming exercises are generated by a teacher LLM and filtered through a validation stage. The resulting datasets are used to adapt smaller models via few-shot prompting, RAG, and LoRA-based fine-tuning. The adapted models are evaluated using benchmark and similarity metrics.}\label{fig:methodology_pipeline}
\end{figure}

This work investigates how smaller open-source language models can be customized for domain-specific code generation using synthetic datasets and different adaptation strategies. Smaller models are attractive because they require fewer computational resources and can be deployed in environments where larger models are impractical. Figure~\ref{fig:methodology_pipeline} illustrates the customization pipeline used in this study.

The proposed pipeline consists of four stages. First, domain-specific programming exercises in the Python ecosystem are generated by an LLM acting as a teacher, where each example pairs a natural language task description with a Python implementation. The teacher model is used only for dataset generation. Second, the generated samples are validated to remove invalid or inconsistent outputs, ensuring that the resulting datasets contain syntactically correct and semantically valid code. Third, the validated datasets are used to adapt smaller language models through few-shot learning, RAG, and LoRA-based fine-tuning. Finally, the adapted models are evaluated using benchmark and similarity metrics. Benchmark evaluations measure functional correctness through automated test cases, while similarity metrics capture how closely generated code aligns with the reference implementations.

The following sections describe each component of the pipeline, including dataset construction, model adaptation strategies, and the evaluation framework.

\section{Synthetic dataset construction}

To support the evaluation of domain-specific code generation models, we construct synthetic datasets of programming exercises using an LLM as a teacher. This follows a knowledge distillation paradigm where a larger model generates training examples later used to adapt smaller models.

In this work, synthetic data is generated using GPT-4o~\cite{openaiGPT4o}, an LLM with strong capabilities in natural language understanding and code generation. The model is used only during dataset creation to produce programming exercises paired with Python implementations. These examples form the basis for subsequent model adaptation experiments.

\subsection{Prompt engineering}

Each example consists of a natural language problem description and a Python implementation. The problem statement appears as a docstring at the beginning of the code snippet, while the solution includes the required imports and commented code implementing the functionality. This structure ensures a consistent format resembling typical programming exercises.

To produce diverse exercises across domains and difficulty levels, we use structured prompts with control variables that guide generation and allow the dataset pipeline to produce tasks with different characteristics. The main variables include:
\begin{itemize}
\item \textbf{topic}: programming concept or library feature targeted by the exercise.
\item \textbf{profession}: domain context simulating a realistic application scenario.
\item \textbf{skill\_level}: difficulty level (e.g., beginner, intermediate, advanced).
\item \textbf{user\_interaction}: whether the exercise requires user interaction.
\item \textbf{error\_handling}: whether the implementation handles invalid inputs.
\end{itemize}

To ensure sufficient topic coverage, an automated pipeline expanded the list of programming topics. For each domain, core topics were identified from official documentation and extended by prompting the teacher model to generate additional subtopics. By sampling combinations of these variables, the pipeline produces exercises covering diverse programming concepts and real-world scenarios.

Figure~\ref{fig:dataset_prompt} shows the prompt template used to generate Python programming exercises. Using this prompt template, the teacher model generates programming exercises that follow a consistent structure. Figure~\ref{fig:dataset_example} shows an example of a generated dataset sample.
\begin{figure}[h]
\centering
\begin{lstlisting}
You are an expert Python instructor tasked with creating specialized programming exercises tailored to various professions. 
Create a Python programming exercise simulating a realistic scenario in the field of {profession},specifically focusing on the topic of {topic}.
The exercise should be suitable for {skill_level} level, include a clear problem statement, be practical and not overly theoretical.
The user interaction should be {user_interaction} and the error handling should be {error_handling}.
The output should be the code with the problem statement as a docstring, and also comments explaining the solution. 
Do not include any further explanations or other text outside the code snippet. Only import conventional libraries. 
The structure of the response should be:
```python
""" 
Problem statement 
"""
imports
code
```
\end{lstlisting}
\caption{\label{fig:dataset_prompt}Prompt template used for dataset generation}
\end{figure}

\begin{figure}[h]
\centering
\begin{lstlisting}[language=Python]
"""
Problem Statement:
In the field of bioinformatics, DNA sequences are often analyzed to find patterns or specific nucleotide occurrences. 
You are given a list of DNA strands. Your task is to write a Python function that counts the occurrence of each 
nucleotide ('A', 'T', 'C', 'G') in a given DNA strand and returns the result as a dictionary.

Instructions:
1. Define a function named `count_nucleotides` that takes a single parameter: a string representing a DNA strand.
2. Initialize a dictionary to store the count of each nucleotide.
3. Iterate over the DNA strand and update the count for each nucleotide in the dictionary.
4. Return the dictionary containing the nucleotide counts.

Example:
Input: "ATCGATCGA"
Output: {'A': 3, 'T': 2, 'C': 2, 'G': 2}
"""

# Define the function to count nucleotides
def count_nucleotides(dna_strand):
    # Initialize a dictionary to store nucleotide counts
    nucleotide_count = {'A': 0, 'T': 0, 'C': 0, 'G': 0}
    
    # Iterate over each character in the DNA strand
    for nucleotide in dna_strand:
        # Update the count for the current nucleotide
        if nucleotide in nucleotide_count:
            nucleotide_count[nucleotide] += 1
    
    # Return the dictionary with counts
    return nucleotide_count

# Example usage
dna_sequence = "ATCGATCGA"
print(count_nucleotides(dna_sequence))  # Output: {'A': 3, 'T': 2, 'C': 2, 'G': 2}
\end{lstlisting}
\caption{\label{fig:dataset_example} Example of a generated programming exercise.}
\end{figure}

 This prompt-based generation pipeline enables the creation of many structured programming exercises for adapting small language models and evaluating domain-specific code generation. The generated samples are then processed through a validation stage to ensure syntactic and semantic correctness.

\subsection{Dataset Analysis}

After generation, the samples were analyzed to understand dataset characteristics and identify potential issues before validation. Using this workflow, approximately $21.6$ thousand exercises were generated per domain. Each request to the teacher model required about $30$ seconds on average, with roughly $150$ input tokens and a maximum generation length of up to $1500$ tokens. Overall, the process produced about $9.7$ million input tokens and $35$ million output tokens, with a total cost of approximately \$374 using the GPT-4o API.

\begin{figure}[h]
\centering
\includegraphics[width=0.9\linewidth]{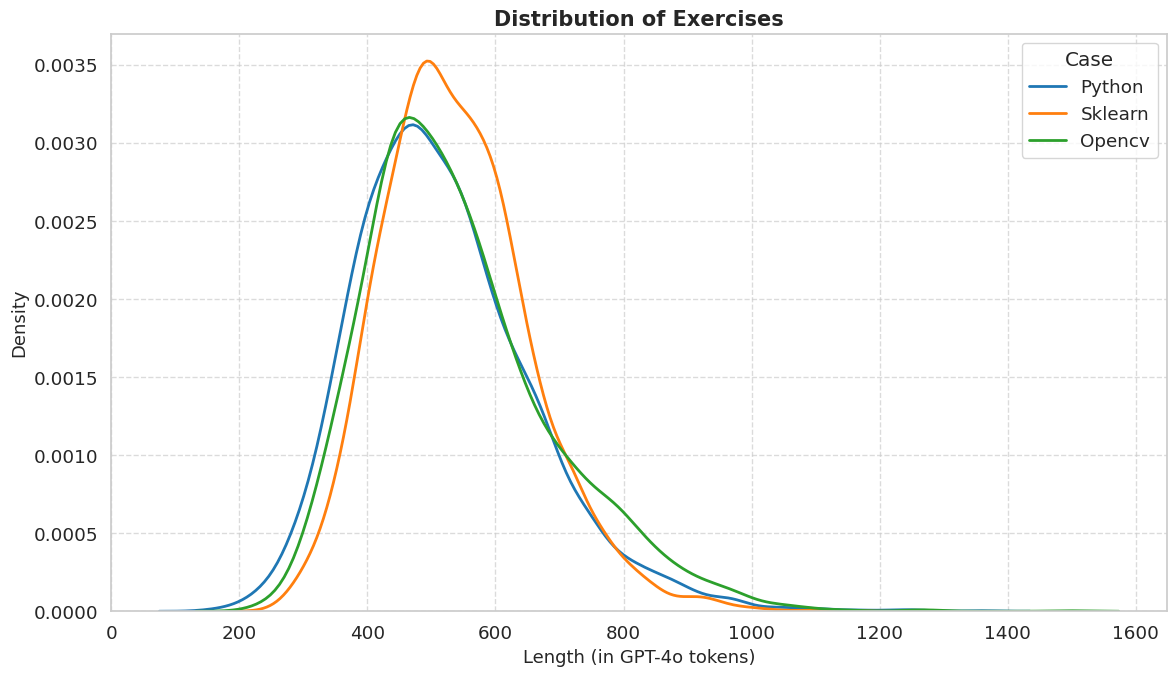}
\caption{Sample length distribution across the three domains.}
\label{fig:length_distribution}
\end{figure}

A subset of approximately $5000$ samples per domain was analyzed to understand the generated dataset. Figure~\ref{fig:length_distribution} shows the distribution of sample lengths, with most samples between 300 and 700 tokens including both the task description and Python implementation. The average length is about 520 tokens for general Python exercises and about 540 tokens for Scikit-learn and OpenCV, indicating similar exercise sizes across domains while preserving sufficient task complexity.

To further examine the structure of the generated exercises, the lengths of the problem descriptions and the corresponding code solutions were analyzed separately. The results show that general Python exercises tend to contain longer textual descriptions, while tasks involving Scikit-learn often produce longer code implementations due to additional steps such as dataset preparation, model training, and evaluation. Exercises in the OpenCV domain exhibit similar behavior, where solutions frequently include multiple image-processing operations.

In addition to length statistics, the generated samples were inspected to verify whether the produced code relies on the expected libraries for each domain. The analysis shows that Python exercises primarily use standard libraries, while Scikit-learn and OpenCV tasks frequently import domain-specific libraries together with supporting packages such as \texttt{numpy}. This indicates that the prompt-based generation process successfully encourages the use of appropriate APIs for each domain.

While the analysis confirms that the generated dataset captures realistic programming patterns, it also reveals occasional inconsistencies in formatting or execution behavior. To ensure dataset reliability, a validation pipeline is applied to filter invalid samples, as described in the following subsection.

\subsection{Dataset Validation}

Although the teacher model can generate coherent programming exercises, the process may occasionally produce malformed or inconsistent outputs, such as incomplete code blocks, syntax errors, missing imports, or references to invalid libraries or API methods. To ensure dataset reliability, generated samples are processed through a two-stage validation pipeline before being used for model adaptation and evaluation.

The first stage verifies the syntactic correctness of the generated code. Each sample is parsed using Python’s abstract syntax tree (AST) parser to ensure valid syntax, and samples that cannot be parsed are removed from the dataset.

The second stage performs semantic validation of imported modules and referenced attributes. Using Python’s module inspection utilities, each import and attribute chain is checked to ensure the referenced APIs exist. Samples referencing invalid modules or attributes are removed.

Table~\ref{tab:dataset_validation} summarizes the validation results. Most samples pass validation, with retention rates above 92\% across all domains. The remaining samples form the final dataset.
\begin{table}[h]
\centering
\caption{Dataset size before and after validation.}\label{tab:dataset_validation}
\begin{tabular}{lccc}
\toprule
Domain & Generated Samples & Valid Samples & Retention Rate \\
\midrule
General Python Exercises & 21,745 & 21,042 & 96.8\% \\
OpenCV Exercises & 22,770 & 21,039 & 92.4\% \\
Scikit-learn Exercises & 20,330 & 20,052 & 98.6\% \\
\bottomrule
\end{tabular}
\end{table}

\subsection{Dataset split}
Validated datasets are split into training, validation, and test sets. The training set is used for fine-tuning, as the document pool for RAG retrieval, and as examples for few-shot prompting. The validation set monitors training performance and computes intermediate metrics, while the test set is reserved for final evaluation.

The datasets are split into $97\%$ training, $1\%$ validation, and $2\%$ test data. This ensures a comparable number of evaluation examples to common code-generation benchmarks, resulting in approximately $200$ validation and $400$ test samples per use case.

\section{Language model customization}

This section describes the models and customization techniques used in our experiments.

\subsection{Base models}

We consider two open-source language models for code generation: \textbf{StarCoder}~\cite{li2023} and \textbf{DeepSeekCoder}~\cite{guo2024}. These models provide strong performance on programming benchmarks while remaining small enough to be adapted with modest computational resources.

\textbf{StarCoder}~\cite{li2023} is a family of decoder-only models ranging from $1$B to $15.5$B parameters. It was trained on \textit{The Stack}~\cite{kocetkov2022}, a large collection of permissively licensed GitHub repositories containing about one trillion tokens across more than $80$ programming languages. The model supports context windows up to $8$k tokens and uses the Fill-in-the-Middle (FIM) objective, enabling both code generation and infilling tasks.

\textbf{DeepSeekCoder}~\cite{guo2024} is a decoder-only code model trained on about two trillion tokens, primarily source code from GitHub repositories across $87$ programming languages. The models range from $1.3$B to $33$B parameters and support context windows up to $16$k tokens. Like StarCoder, DeepSeekCoder uses the FIM objective and supports code completion and infilling.

In contrast, GPT-4o~\cite{openaiGPT4o} is used only as a \textit{teacher model} during dataset generation. Due to its large scale and proprietary nature, it is not customized in our experiments. Instead, it generates synthetic training data used to adapt the smaller open-source models to the target domains.

\subsection{Customization techniques}

We investigate three strategies for adapting language models to domain-specific code generation: few-shot learning, retrieval-augmented generation (RAG), and parameter-efficient fine-tuning.

\textbf{Few-shot learning} guides the model during inference by providing example task–solution pairs in the prompt context. This approach leverages the in-context learning capabilities of language models without modifying their parameters~\cite{brown2020}. In our setup, examples from the synthetic training dataset are appended to the prompt to illustrate how similar tasks should be solved.

\textbf{Retrieval-Augmented Generation (RAG)} enhances prompting by dynamically retrieving relevant examples from an external knowledge source~\cite{gao2024}. In our setup, training examples are embedded using the \textit{sentence-transformers/all-MiniLM-L6-v2}~\cite{allMiniLM2025} model and stored in a vector database. During inference, the query embedding retrieves the most similar examples, which are inserted into the prompt. This extends the model’s effective memory and enables access to domain-specific examples beyond the static prompt.

\textbf{Fine-tuning} specializes the model by updating parameters using the synthetic dataset. Instead of full fine-tuning, we use \textit{Low-Rank Adaptation (LoRA)}~\cite{hu2022}, a parameter-efficient method that freezes the original weights and introduces trainable low-rank matrices to approximate weight updates. This reduces the number of trainable parameters while preserving the base model’s knowledge. LoRA is applied to the attention projection matrices of the Transformer architecture following the standard PEFT setup~\cite{han2024}.

\subsection{Inference and evaluation setup}

All experiments use a consistent inference configuration across customization techniques. Code generation uses greedy decoding, selecting the most probable token at each step to ensure reproducibility. Model performance is evaluated using both benchmark-based and similarity metrics.

\textbf{Benchmark evaluation} measures the functional correctness of the code generated using automated test suites. For general Python tasks we use the HumanEval benchmark~\cite{chen2021}, which contains $164$ programming problems with unit tests. For domain-specific tasks we use subsets of BigBenchCode~\cite{zhuo2025} for the Scikit-learn and OpenCV libraries (\textit{BCSk} with 152 tasks and \textit{BCCV} with 10 tasks).

While benchmarks measure functional correctness, they do not capture whether generated code aligns with domain-specific conventions. We therefore also consider \textbf{similarity evaluation}, which measures how closely generated code matches the target dataset. Cosine similarity is computed between embeddings of generated solutions and reference implementations using the same embedding model employed for retrieval. Two variants are used: validation similarity (to guide fine-tuning) and test similarity (for final evaluation). Together, these metrics provide a complementary view of model performance, capturing both functional correctness and structural similarity to domain-specific code.

\section{Results}

We present the empirical evaluation of the customization techniques. We report baseline performance and analyze the impact of few-shot learning, RAG, and LoRA-based fine-tuning. Performance is measured using benchmark-based metrics (Pass@1) and similarity metrics capturing alignment with the target datasets. For general Python tasks we use the HumanEval benchmark, while for library-specific tasks we use BigBenchCode subsets: BCSk for Scikit-learn and BCCV for OpenCV.

Table~\ref{tab:results_summary} summarizes the results. For each customization technique, we report benchmark accuracy (Pass@1), similarity scores, and the change relative to the baseline. A detailed hyperparameter analysis is provided in~\cite{freire2025}.
\begin{table}[t]
\centering
\small
\caption{Performance of customization techniques across domains. Pass@1 measures benchmark correctness, and Sim. denotes cosine similarity to reference solutions. Values in parentheses indicate changes relative to the baseline (percentage points). Few-shot and RAG use 3 examples, with RAG applying a threshold of 0.5. LoRA fine-tuning uses $r=128$ and $\alpha=128$. Bold values indicate the best result per model, while the blue background highlights the best result across models.}\label{tab:results_summary}
\begin{tabular}{l
S@{\hspace{2pt}}c S@{\hspace{2pt}}c
@{\hspace{10pt}}
S@{\hspace{2pt}}c S@{\hspace{2pt}}c
@{\hspace{10pt}}
S@{\hspace{2pt}}c S@{\hspace{2pt}}c}

\toprule
& \multicolumn{4}{c}{Python}
& \multicolumn{4}{c}{Scikit-learn}
& \multicolumn{4}{c}{OpenCV} \\
\cmidrule(lr){2-5}
\cmidrule(lr){6-9}
\cmidrule(lr){10-13}

Method
& \multicolumn{2}{c}{Pass@1} & \multicolumn{2}{c}{Sim.}
& \multicolumn{2}{c}{Pass@1} & \multicolumn{2}{c}{Sim.}
& \multicolumn{2}{c}{Pass@1} & \multicolumn{2}{c}{Sim.} \\

\midrule

\multicolumn{13}{c}{\textbf{StarCoder-1B}} \\
\midrule

Baseline
& 16.0 & {} & 73.6 & {}
& 13.2 & {} & 62.8 & {}
& 0.0 & {} & 79.4 & {} \\

Few-shot
& 14.6 & {\color{darkred}(-1.4)} 
& 75.8 & {\color{darkgreen}(+2.2)}
& 3.9 & {\color{darkred}(-9.3)} 
& 68.0 & {\color{darkgreen}(+5.2)}
& 10.0 & {\color{darkgreen}(+10.0)} 
& 81.1 & {\color{darkgreen}(+1.7)} \\

RAG
& 14.0 & {\color{darkred}(-2.0)} 
& 80.9 & {\color{darkgreen}(+7.3)}
& 4.6 & {\color{darkred}(-8.6)} 
& 74.5 & {\color{darkgreen}(+11.7)}
& 0.0 & {}
& 87.6 & {\color{darkgreen}(+8.2)} \\

LoRA
& \textbf{18.3} & {\color{darkgreen}(+2.3)} 
& \textbf{87.1} & {\color{darkgreen}(+13.5)}
& \textbf{20.4} & {\color{darkgreen}(+7.2)} 
& \textbf{77.5} & {\color{darkgreen}(+14.7)}
& \textbf{20.0} & {\color{darkgreen}(+20.0)} 
& \textbf{87.1} & {\color{darkgreen}(+7.8)} \\

\midrule

\multicolumn{13}{c}{\textbf{DeepSeekCoder-1.3B}} \\
\midrule

Baseline
& 30.5 & {} & 84.2 & {}
& 32.9 & {} & 73.2 & {}
& 20.0 & {} & 85.9 & {} \\

Few-shot
& 39.0 & {\color{darkgreen}(+8.5)} 
& 82.2 & {\color{darkred}(-2.0)}
& 28.9 & {\color{darkred}(-4.0)}   
& 74.5 & {\color{darkgreen}(+1.3)}
& 0.0 & {\color{darkred}(-20.0)}  
& 86.4 & {\color{darkgreen}(+0.5)} \\

RAG
& \cellcolor{best}\textbf{39.6} & {\color{darkgreen}(+9.1)}
& 85.6 & {\color{darkgreen}(+1.4)}
& 29.6 & {\color{darkred}(-3.3)}
& 78.5 & {\color{darkgreen}(+5.3)}
& 20.0 & 
& 88.2 & {\color{darkgreen}(+2.3)} \\

LoRA
& 38.4 & {\color{darkgreen}(+7.9)}
& \cellcolor{best}\textbf{88.2} & {\color{darkgreen}(+4.0)}
& \cellcolor{best}\textbf{33.9} & {\color{darkgreen}(+1.0)}
& \cellcolor{best}\textbf{81.9} & {\color{darkgreen}(+9.4)}
& \cellcolor{best}\textbf{50.0} & {\color{darkgreen}(+30.0)}
& \cellcolor{best}\textbf{90.9} & {\color{darkgreen}(+6.0)} \\

\bottomrule
\end{tabular}
\end{table}

\subsection{Baseline performance}

Before applying customization, we evaluate the base models on the benchmarks to establish a reference point.

Overall, DeepSeekCoder-1.3B outperforms StarCoder-1B in the baseline setting. On the Python benchmark, DeepSeekCoder achieves $30.5\%$ Pass@1 compared to $16.0\%$ for StarCoder. A similar pattern appears in the domain-specific benchmarks. In particular, DeepSeekCoder reaches $20.0\%$ Pass@1 on OpenCV, while StarCoder fails to solve any tasks.

The similarity metrics follow the same pattern, with DeepSeekCoder producing outputs closer to the reference implementations. This indicates that its stronger baseline capabilities provide a better starting point for domain-specific code generation.

At the same time, the lower scores on the Scikit-learn and OpenCV benchmarks highlight the difficulty of generating code that correctly uses specialized APIs and library conventions. This motivates the use of customization techniques to improve domain alignment and functional correctness.

\subsection{Few-shot learning}
Few-shot learning is evaluated by providing different numbers of example task–solution pairs in the prompt. We test configurations ranging from one to ten examples, depending on the model’s context length. In Table~\ref{tab:results_summary}, we report the 3-shot results for simplicity.

The results show that few-shot prompting can slightly increase similarity to the target dataset, particularly with three to five prompt examples. However, improvements in benchmark accuracy are limited. In some cases, adding more examples introduces noise or reduces performance due to context window limits.

Overall, few-shot learning provides modest improvements with minimal implementation cost but remains limited by prompt length and the model’s existing knowledge. These limitations motivate retrieval-augmented generation, which provides relevant context dynamically instead of relying on static prompt examples.

\subsection{Retrieval-Augmented Generation}

RAG extends few-shot prompting by dynamically retrieving relevant examples instead of relying on a fixed set of prompt examples. In our setup, training examples are embedded and stored in a vector database, and the most similar ones are retrieved at inference time and inserted into the prompt.

Similar to the few-shot experiments, we evaluate different numbers of retrieved examples and similarity thresholds for filtering retrieved documents. In Table~\ref{tab:results_summary}, we report results using 3 retrieved examples with a threshold of $0.5$.

The results show that RAG can substantially increase similarity to the target dataset, indicating that retrieved examples help guide the model toward domain-specific coding patterns. This effect is particularly visible for StarCoder-1B, with similarity increases of up to $+11.7$ points for Scikit-learn and $+8.2$ points for OpenCV.

However, improvements in benchmark accuracy are less consistent. DeepSeekCoder achieves stronger Python performance with RAG ($39.6\%$ Pass@1), but the approach can reduce performance for StarCoder and on library-specific tasks. This suggests that retrieved examples may improve stylistic alignment with the dataset without consistently improving functional correctness.

Overall, RAG improves domain alignment more consistently than few-shot prompting, but its impact on benchmark accuracy depends on the quality of the retrieved examples.

\subsection{Fine-tuning}

We evaluate parameter-efficient fine-tuning using LoRA. The base models are adapted using the synthetic training dataset while keeping the original model weights frozen and training only the low-rank adaptation matrices.

\begin{figure}[t]
\centering
\includegraphics[width=0.8\linewidth]{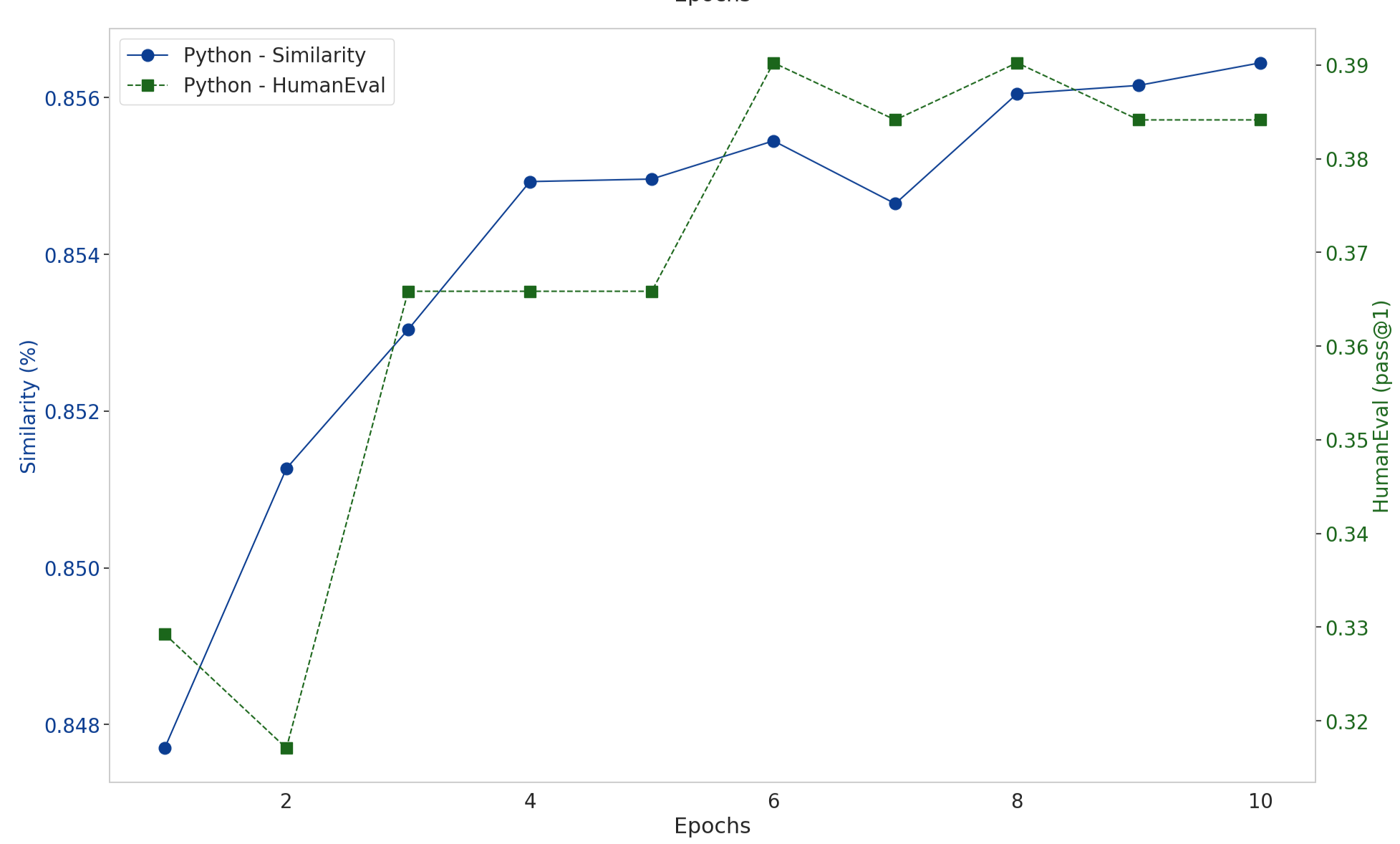}
\caption{Training dynamics during LoRA fine-tuning for DeepSeekCoder on Python tasks ($\alpha = r = 128$). The plot shows validation similarity and HumanEval Pass@1 during training. Increases in validation similarity correlate with improvements in benchmark accuracy.}
\label{fig:training_dynamics}
\end{figure}

During fine-tuning, we monitor both validation similarity and benchmark performance. Figure~\ref{fig:training_dynamics} illustrates the evolution of these metrics during LoRA training for the Python tasks. In general, improvements in validation similarity correlate with increases in benchmark accuracy, supporting its use as a signal for selecting the final model.

Several fine-tuning configurations were explored, including different training settings and LoRA hyperparameters. For brevity, Table~\ref{tab:results_summary} reports results for $r=128$ and $\alpha=128$. This corresponds to $57.4$M trainable parameters for StarCoder-1B and $50.4$M for DeepSeekCoder-1.3B.

LoRA produces the largest gains among the customization techniques. Compared to the baseline, fine-tuning consistently improves both benchmark accuracy and similarity to the target datasets. Gains are particularly pronounced for the domain-specific benchmarks. On OpenCV tasks, StarCoder gains $+20.0$ Pass@1 points and DeepSeekCoder $+30.0$. StarCoder also gains $+7.2$ points on the Scikit-learn benchmark.

Fine-tuning also yields the largest similarity gains across domains, indicating that the models learn coding patterns closer to the target datasets. For example, StarCoder gains $+14.7$ similarity points on Scikit-learn tasks, while DeepSeekCoder gains $+9.4$.

Overall, these results show that LoRA-based fine-tuning is the most effective customization strategy in our experiments, yielding substantial gains in both functional correctness and domain alignment compared to prompting-based approaches.

\subsection{Comparison of customization techniques}

The three customization strategies differ not only in performance but also in data requirements, computational cost, and deployment complexity.

\textbf{Few-shot learning} is the simplest approach, requiring no additional training or infrastructure. By inserting a small number of example task–solution pairs into the prompt, the model leverages in-context learning to adapt its output. However, this approach is limited by the model’s context window and relies on knowledge encoded during pre-training. In our experiments, few-shot prompting yields only modest improvements and does not consistently improve benchmark accuracy.

\textbf{RAG} extends this idea by dynamically retrieving relevant examples instead of relying on a fixed set of prompt examples. This allows the model to access more domain-specific examples than can fit in the prompt. In our experiments, RAG consistently increases similarity to the target datasets, indicating that retrieved examples guide the model toward domain-specific coding patterns. However, its impact on benchmark accuracy is less consistent, as retrieved examples may introduce unnecessary operations that reduce functional correctness.

\textbf{Fine-tuning} requires the largest upfront investment in data preparation and computational resources. Training LoRA adapters updates millions of parameters and requires GPU resources. However, once trained, the adapted model can be deployed without long prompts or retrieval infrastructure. In our experiments, LoRA-based fine-tuning yields the most consistent improvements in both benchmark accuracy and similarity, particularly for domain-specific tasks.

The effectiveness of each customization strategy also depends on the base model. DeepSeekCoder generally benefits more from in-context methods such as few-shot learning and RAG, likely due to its stronger baseline and larger pre-training dataset. In contrast, StarCoder shows larger gains from fine-tuning, suggesting that parameter updates better compensate for limitations in its pre-training.

Figure~\ref{fig:python_qualitative_example} illustrates these differences with example solutions for a HumanEval task. The baseline model produces an incorrect recursive implementation due to a missing base case, while the customized approaches generate valid solutions. In particular, the fine-tuned model produces a concise implementation that follows common coding patterns.
\begin{figure}[h]
\centering

\begin{subfigure}{0.48\linewidth}
\textbf{Baseline - Fail}

\begin{minipage}[t][5cm][t]{\linewidth}
\begin{lstlisting}[language=Python]
if n == 0:
    return 0
elif n == 1:
    return 0
elif n == 2:
    return 2
else:
    return fib4(n - 1) + fib4(n - 2)
        + fib4(n - 3) + fib4(n - 4)
\end{lstlisting}
\end{minipage}
\end{subfigure}
\hfill
\begin{subfigure}{0.48\linewidth}
\textbf{Few-shot - Pass}

\begin{minipage}[t][5cm][t]{\linewidth}
\begin{lstlisting}[language=Python]
# Initialize the first two Fib4 numbers
fib4_numbers = [0, 0, 2, 0]

# Iterate through the remaining Fib4 numbers
for i in range(4, n + 1):
    # Calculate the next Fib4 number
    fib4_numbers.append(
        fib4_numbers[i - 1] +
        fib4_numbers[i - 2] +
        fib4_numbers[i - 3] +
        fib4_numbers[i - 4]
    )

# Return the n-th Fib4 number
return fib4_numbers[n]
\end{lstlisting}
\end{minipage}
\end{subfigure}

\begin{subfigure}{0.48\linewidth}
\textbf{RAG - Pass}

\begin{minipage}[t][4.5cm][t]{\linewidth}
\begin{lstlisting}[language=Python]
# Base case
if n == 0:
    return 0
elif n == 1:
    return 0
elif n == 2:
    return 2
elif n == 3:
    return 0

# Recursive case
return fib4(n - 1) + fib4(n - 2) + fib4(n - 3) + fib4(n - 4)
\end{lstlisting}
\end{minipage}
\end{subfigure}
\hfill
\begin{subfigure}{0.48\linewidth}
\textbf{LoRA Fine-tuned - Pass}

\begin{minipage}[t][4.5cm][t]{\linewidth}
\begin{lstlisting}[language=Python]
# Initialize the list of Fib4 numbers
fib4_numbers = [0, 0, 2, 0]

# Iterate through the sequence up to n
for i in range(4, n + 1):
    # Calculate the next Fib4 number using
    # the previous 4 numbers
    fib4_numbers.append(
        sum(fib4_numbers[-4:])
    )

# Return the n-th Fib4 number
return fib4_numbers[n]
\end{lstlisting}
\end{minipage}
\end{subfigure}

\caption{Example solutions for HumanEval task 46 across customization strategies.}
\label{fig:python_qualitative_example}
\end{figure}

Overall, while prompting-based approaches offer simplicity and low deployment cost, LoRA-based fine-tuning provides the most reliable strategy for adapting smaller language models to specialized domains, yielding the largest gains in both functional correctness and domain alignment.

\section{Conclusion}

We investigated how smaller open-source language models can be customized for domain-specific code generation using synthetic datasets and lightweight adaptation techniques. Our pipeline uses a large language model to generate synthetic programming exercises that are then used to adapt smaller models through prompting, retrieval, and parameter-efficient fine-tuning.

Experiments across three domains—general Python, Scikit-learn, and OpenCV—show that customization improves the performance of smaller models. Few-shot learning provides a simple and low-cost mechanism but yields modest improvements. Retrieval-augmented generation increases similarity to the target datasets by providing relevant examples at inference time, although its impact on benchmark accuracy is less consistent. Among the evaluated approaches, LoRA-based fine-tuning achieves the largest gains in both functional correctness and domain alignment.

The results also highlight the role of the base model. DeepSeekCoder performs better in the baseline setting and benefits more from in-context methods, while StarCoder shows larger gains when fine-tuned. This suggests that parameter-efficient fine-tuning can compensate for limitations in the pre-training of smaller models.

Overall, these findings show that combining synthetic dataset generation with lightweight customization techniques provides a practical strategy for adapting smaller language models to specialized programming domains. Future work may extend this pipeline to additional domains, improve retrieval strategies for RAG, and develop better evaluation methods for domain-specific code generation.

\bibliographystyle{unsrtnat}

\bibliography{references}

@STRING{JMLR        =   "Journal of Machine Learning Research"}

@STRING{ACL         =   "Annual Meeting of the Association for Computational Linguistics"}

@STRING{ICLR        = "International Conference on Learning Representations"}

@STRING{NEURIPS     =   "Advances in Neural Information Processing Systems"}

@STRING{ARXIV       =   "{arXiv} preprint"}

@misc{allMiniLM2025,
    author          =   "Sentence-Transformers",
    title           =   "{all-MiniLM-L6-v2} model on {HuggingFace}",
    url             =   "https://huggingface.co/sentence-transformers/all-MiniLM-L6-v2",
    note            =   "Accessed: 2025-08-20"
}

@inproceedings{brown2020,
    title           =   "Language Models are Few-Shot Learners",
    author          =   "Tom B. Brown and Benjamin Mann and Nick Ryder and Melanie Subbiah and Jared Kaplan and Sam McCandlish and Greg Brockman and OpenAI Team",
    booktitle       =   NEURIPS,
    pages           =   "1877--1901",
    year            =   2020
}

@inproceedings{chen2021,
    title           =   "Evaluating Large Language Models Trained on Code",
    author          =   "Mark Chen and Jerry Tworek and others",
    booktitle       =   ICLR,
    pages           =   "1--13",
    year            =   2022
}

@article{eldan2023,
    title           =   "{TinyStories}: How Small Can Language Models Be and Still Speak Coherent English?", 
    author          =   "Ronen Eldan and Yuanzhi Li",
    journal         =   ARXIV,
    volume          =   "{arXiv}:2305.07759",
    year            =   2023
}

@mastersthesis{freire2025,
    author          =   "Luis Freire",
    title           =   "Exploring different approaches to customize {LLM}s for text-to-code generation",
    school          =  "Technical University of Denmark",
    type            =   "Master's thesis",
    year            =   2025
}

@article{gao2024,
    author          =   "Yunfan Gao and Yun Xiong and Xinyu Gao and Kangxiang Jia and Jinliu Pan and Yuxi Bi and Yi Dai and Jiawei Sun and Meng Wang and Haofen Wang",
    title           =   "{Retrieval-Augmented} {Generation} for Large Language Models: A Survey",
    journal         =   ARXIV,    
    volume          =   "{arXiv}:2312.10997",
    year            =   2024
}

@article{gunasekar2023,
    title           =   "Textbooks Are All You Need", 
    author          =   "Suriya Gunasekar and Yi Zhang and Jyoti Aneja and Caio César Teodoro Mendes and Allie Del Giorno and Sivakanth Gopi and Mojan Javaheripi and Piero Kauffmann and Gustavo de Rosa and Olli Saarikivi and Adil Salim and Shital Shah and Harkirat Singh Behl and Xin Wang and Sébastien Bubeck and Ronen Eldan and Adam Tauman Kalai and Yin Tat Lee and Yuanzhi Li",
    journal         =   ARXIV,
    volume          =   "{arXiv}:2306.11644",
    year            =   2023
}

@article{guo2024,
    title           =   "{DeepSeek-Coder}: When the Large Language Model Meets Programming -- The Rise of Code Intelligence", 
    author          =   "Daya Guo and Qihao Zhu and Dejian Yang and Zhenda Xie and Kai Dong and Wentao Zhang and Guanting Chen and Xiao Bi and Y. Wu and Y. K. Li and Fuli Luo and Yingfei Xiong and Wenfeng Liang",
    journal         =   ARXIV,
    volume          =   "{arXiv}:2401.14196",
    year            =   2024
}

@article{han2024,
    title           =   "{Parameter}-{Efficient} {Fine-Tuning} for Large Models: A Comprehensive Survey", 
    author          =   "Zeyu Han and Chao Gao and Jinyang Liu and Jeff Zhang and Sai Qian Zhang",
    journal         =   ARXIV,
    volume          =   "{arXiv}:2403.14608",
    year            =   2024
}

@article{hinton2015,
    title           =   "Distilling the Knowledge in a Neural Network",
    author          =   "Hinton, Geoffrey and Vinyals, Oriol and Dean, Jeff",
    journal         =   ARXIV,
    volume          =   "{arXiv}:1503.02531",
    year            =   2015
}

@inproceedings{hu2022,
    title           =   "{LoRA}: Low-Rank Adaptation of Large Language Models",
    author          =   "Hu, Edward J and Shen, Yelong and Wallis, Phillip and Allen-Zhu, Zeyuan and Li, Yuanzhi and Wang, Shean and Wang, Lu and Chen, Weizhu",
    booktitle       =   ICLR,
    pages           =   "1--10",
    year            =   2022
}

@article{kaplan2020,
    title           =   "Scaling Laws for Neural Language Models", 
    author          =   "Jared Kaplan and Sam McCandlish and Tom Henighan and Tom B. Brown and Benjamin Chess and Rewon Child and Scott Gray and Alec Radford and Jeffrey Wu and Dario Amodei",
    journal         =   ARXIV,
    volume          =   "{arXiv}:2001.08361",
    year            =   2020
}

@article{kocetkov2022,
    title           =   "The {Stack}: 3 {TB} of permissively licensed source code", 
    author          =   "Denis Kocetkov and Raymond Li and Loubna Ben Allal and Jia Li and Chenghao Mou and Carlos Muñoz Ferrandis and Yacine Jernite and Margaret Mitchell and Sean Hughes and Thomas Wolf and Dzmitry Bahdanau and Leandro von Werra and Harm de Vries",
    journal         =   ARXIV,
    volume          =   "{arXiv}:2211.15533",
    year            =   2022
}

@article{li2023,
    title           =   "{StarCoder}: may the source be with you!", 
    author          =   "Raymond Li and Loubna Ben Allal and others",
    journal         =   ARXIV,
    volume          =   "{arXiv}:2305.06161",
    year            =   2023
}

@inproceedings{lin2004,
    title           =   "{ROUGE}: A Package for Automatic Evaluation of Summaries",
    author          =   "Lin, Chin-Yew",
    booktitle       =   ACL,
    pages           =   "74--81",
    year            =   2004
}

@article{minaee2025,
    title           =   "Large Language Models: A Survey", 
    author          =   "Shervin Minaee and Tomas Mikolov and Narjes Nikzad and Meysam Chenaghlu and Richard Socher and Xavier Amatriain and Jianfeng Gao",
    journal         =   ARXIV,
    volume          =   "{arXiv}:2402.06196",
    year            =   2025
}

@misc{openaiGPT4o,
    author          =   "{OpenAI}",
    title           =   "Hello {GPT}-4o",
    url             =   "https://openai.com/index/hello-gpt-4o/",
    year            =   2024,
    note            =   "Accessed: 2025-08-15"
}

@article{opencv_library,
    author          =   "Gary Bradski",
    title           =   "{The OpenCV Library}",
    journal         =   "{Dr. Dobb}'s Journal of Software Tools",
    year            =   2000
}

@inproceedings{papineni2002,
    title           =   "{Bleu}: a Method for Automatic Evaluation of Machine Translation",
    author          =   "Papineni, Kishore and Roukos, Salim and Ward, Todd and Zhu, Wei-Jing",
    booktitle       =   ACL,
    pages           =   "311--318",
    year            =   "2002"
    
}

@article{ren2020,
    author          =   "Shuo Ren and Daya Guo and Shuai Lu and Long Zhou and Shujie Liu and Duyu Tang and Neel Sundaresan and Ming Zhou and Ambrosio Blanco and Shuai Ma",
    title           =   "{CodeBLEU}: a Method for Automatic Evaluation of Code Synthesis",
    journal         =   ARXIV,
    volume          =   "{arXiv}:2009.10297",
    year            =   2020
}

@article{scikit-learn,
    title           =   "{Scikit-learn}: Machine Learning in {P}ython",
    author          =   "Fabian Pedregosa and Gaël Varoquaux and Alexandre Gramfort and Vincent Michel and Bertrand Thirion and others",
    journal         =   JMLR,
    volume          =   12,
    pages           =   "2825--2830",
    year            =   2011
}

@inproceedings{vaswani2017,
    title           =   "Attention Is All You Need",
    author          =   "Vaswani, Ashish and Shazeer, Noam and Parmar, Niki and Uszkoreit, Jakob and Jones, Llion and Gomez, Aidan N. and Kaiser, Lukasz and Polosukhin, Illia",
    booktitle       =   NEURIPS,
    pages           =   "6000--6010",
    year            =   2017
}

@article{zhuo2025,
    title           =   "{BigCodeBench}: Benchmarking Code Generation with Diverse Function Calls and Complex Instructions", 
    author          =   "Terry Yue Zhuo and Minh Chien Vu and others",
    journal         =   ARXIV,
    volume          =   "{arXiv}:2406.15877",
    year            =   2025
}


\end{document}